\documentclass[letter, conference, final]{IEEEtran}

\makeatletter
\def\bstctlcite{\@ifnextchar[{\@bstctlcite}{\@bstctlcite[@auxout]}}
\def\@bstctlcite[#1]#2{\@bsphack
  \@for\@citeb:=#2\do{%
    \edef\@citeb{\expandafter\@firstofone\@citeb}%
    \if@filesw\immediate\write\csname #1\endcsname{\string\citation{\@citeb}}\fi}%
  \@esphack}
\makeatother

\usepackage[vlined, ruled, linesnumbered, commentsnumbered]{algorithm2e}
\usepackage{lipsum}
\usepackage{multicol}
\usepackage{multirow}
\usepackage{amsmath}
\usepackage{cite}
\usepackage{color}
\usepackage{xcolor}
\definecolor{chred}{rgb}{0.8,0,0}
\definecolor{chgray}{rgb}{0.5,0.5,0.5}
\usepackage{graphicx}
\usepackage{caption}
\usepackage{subcaption}
\usepackage{dblfloatfix}
\usepackage{eqlist}
\usepackage{txfonts}
\usepackage{url}
\usepackage{footmisc}
\usepackage{booktabs}
\usepackage{balance}

\usepackage{multirow}
\usepackage{threeparttable}
\begin{document}
\IEEEoverridecommandlockouts

\title{\LARGE \bf
A Hand Combining Two Simple Grippers to Pick up and Arrange Objects for
Assembly}

\author{Kaidi Nie$^1$, Weiwei Wan$^{1,2,*}$, and Kensuke Harada$^{1,2}$
 \thanks{$^{1}$Graduate School of Engineering Science, Osaka University, Japan.
$^{2}$National Inst. of AIST. *Correspondance author: Weiwei Wan {\tt\small
wan@hlab.sys.es.osaka-u.ac.jp}}%
}
\maketitle
\thispagestyle{empty}
\pagestyle{empty}

\begin{abstract}
This paper proposes a novel robotic hand design for assembly tasks. The idea is
to combine two simple grippers -- an inner gripper which is used for precise
alignment, and an outer gripper which is used for stable holding.
Conventional robotic hands require complicated compliant mechanisms or
complicated control strategy and force sensing to conduct assemble tasks, which
makes them costly and difficult to pick and arrange small objects like screws or
washers. Compared to the conventional hands, the proposed design provides a
low-cost solution for aligning, picking up, and arranging various objects by
taking advantages of the geometric constraints of the positioning fingers and gravity.
It is able to deal with small screws and washers, and eliminate
the position errors of cylindrical objects or objects with cylindrical holes.
In the experiments, both real-world tasks and quantitative analysis are
performed to validate the aligning, picking, and arrangements abilities of
the design.
\end{abstract}


\section{Introduction}

The goal of this paper is to develop a robotic hand to pick up and arrange
objects for assembly. Assembly is a classical problem to the robotics community.
Researchers have studied the problem for decades and developed several solutions like
compliant mechanisms and sensor-based close-loop control to perform assembly
tasks. Using modern force sensors or well designed compliant mechanisms,
assembly tasks could be performed with a high success rate.

On the other hand, the available solutions are relatively complicated and
costly. They require high-quality force sensors and delicately designed
mechanisms which might be unaffordable for middle and small manufacturers.
Besides, especially for small objects such as nuts, washers, and screws which are not easy to grip,
extra tools are needed. The extra tools impair the flexibility of an assembly
system. For these reasons, we propose a novel hand design in this paper by
combining an internal gripper, which is used to align objects and
eliminate position errors, and an outer gripper, which is used for stable
holding.

\begin{figure}[!htbp]
\begin{center}
\includegraphics[width=0.45\textwidth]{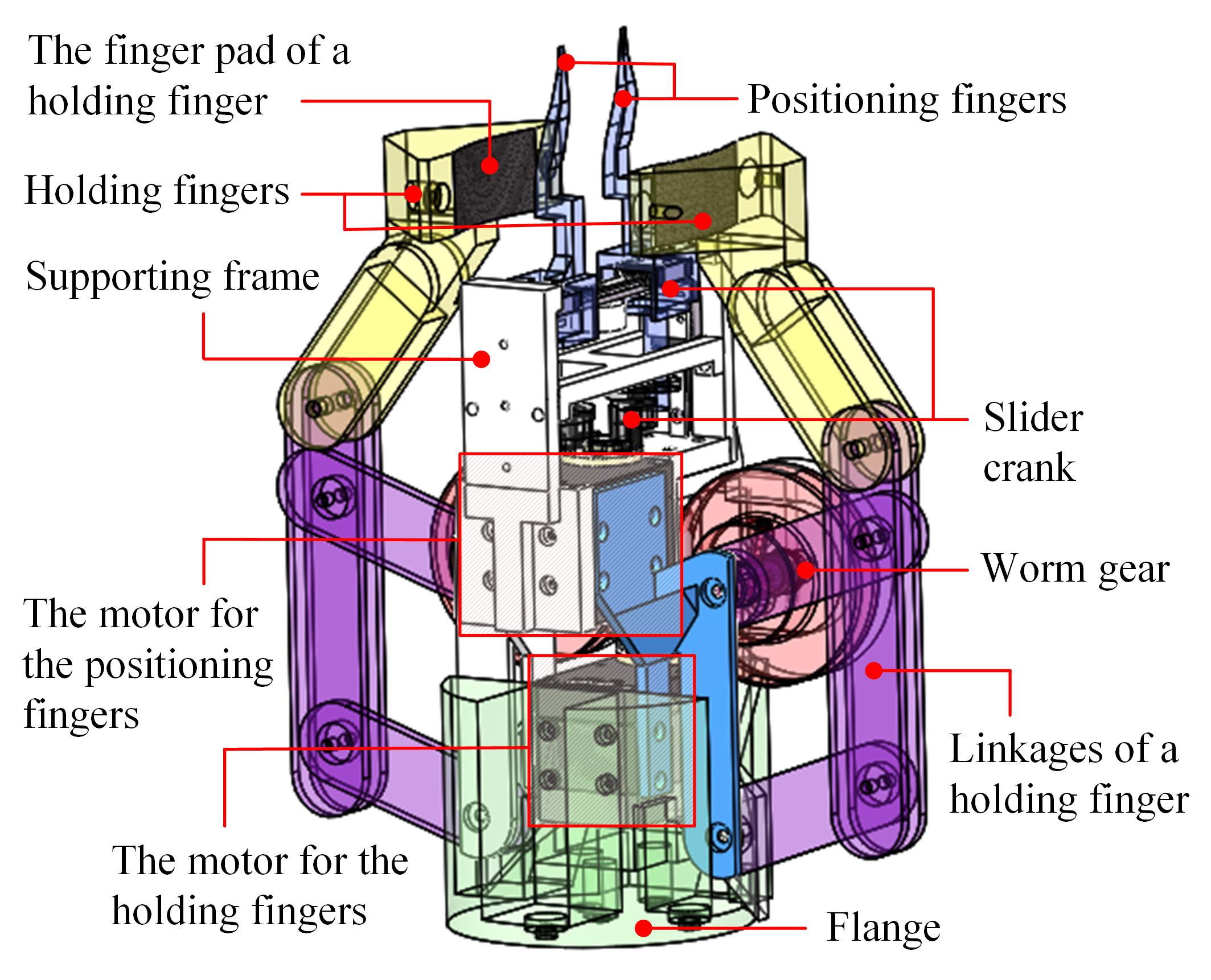}
\caption{A sketch of the proposed design. It comprises two simple
grippers -- an inner gripper with positioning fingers for object alignment, and
an outer gripper with holding fingers to grasp objects stably.
The positioning fingers are driven by a slider-crank mechanism.
The holding fingers are driven by a parallelogram actuated by
a worm set. The blue part is the mainframe. It connects both the worm gears and
the two motors to strength the structure of the hand. The green part at
the bottom is a flange used to connect to robot arms.}
\label{teaser}
\end{center}
\end{figure}

The goal of our design is as follows: 1) The gripper is expected to be able
to generate large forces and hold objects stably. 2) The gripper is expected to
have the ability to reduce the position errors of cylindrical objects or
objects with cylindrical holes. 3) The gripper is expected to be able to pick
and arrange small objects such as nuts, washers, and screws. 4) The gripper is
expected to be simple in both mechanism and control. The proposed hand design
satisfies these four requirements by combining two simple grippers.

Fig.\ref{teaser} shows a sketch of the proposed design. 
In the inner part, there is a linear gripper driven by a slider-crank mechanism.
The fingers of this gripper are thin and deliberately cut to align and
arrange objects. In the outer part, there is a rotational gripper driven by a
parallelogram. The gripper has two big fingers to exert large holding force. 
The rotational design of the holding gripper enables it to fold the fingers
back when they are not used, and avoid conflict with the positioning gripper.

The positioning fingers could be used to eliminate the position errors of
cylindrical objects or objects with cylindrical holes. It may grip cylindrical
objects by closing the two fingers externally or grip objects with cylindrical
holes by stretching the two fingers internally from the holes. 
The positioning fingers could also be used to pick up small objects like nuts
and washers, and arrange screws by taking advantages of the geometric
constraints from their delicately cut surfaces and gravity. The holding
fingers are used to grasp objects stably after alignment.

In the experimental section, different objects and various tests are used to
validate the aligning, picking, and arrangements abilities of the design. The
holding forces of the outer gripper, and the aligning and arranging abilities
of the inner gripper, are analyzed qualitatively, and the performance of a
prototype is demonstrated with several real-world tasks.

The organization of the paper is as follow. Related work is presented in Section
II. Details of the design are shown in Section III. Section IV
explains the working strategies. Experiments and analysis are
performed in Section V. Conclusions are drawn in Section VI.

\section{Related Work}

The goal of this paper is to develop a simple robotic hand to align, pick
up, and arrange objects for assembly. Thus, we review the
literature related to gripper-based robotic hand design \cite{robotgripper}.
We present the related publications that have similar design goals or
implemented similar functions.

For the work with similar design goals, Nishimura et al.
\cite{nishimura2018lightweight} designed a gripper that can position and hold
objects using a chuck clamping mechanism. The surface of their fingertips is
carefully cut to generate tangential contact forces and push target objects to
the hand center. Chen et al. \cite{chen2014hand} designed a gripper that can
twist and re-positioning objects by combining linear motors and rotational motors. 
Hirata et al. \cite{hirata2011design} presented the design of fingertips
to cage, align, and firmly pick up small circular parts like bearings, washers,
and gears. The design is later used in real-world assembly tasks in
\cite{yamaguchi2012design}. More generally, Alberto et al.
\cite{rodriguez2013effector} proposed the design of finger shapes for 1DoF
planar actuation, where he proposed a tool to transform the geometry contact
constraints into an effector shape.
Hsu et al. \cite{hsu2017self} used a self-locking underactuated mechanism
mounted in parallel to actuators to firmly grasp objects. The self-locking is
triggered automatically when the desired grasp is achieved. Without the
self-locking mechanism, their hand design is similar to
\cite{odhner2014compliant}. 
Harada et al. \cite{harada2016proposal} proposed a gripper design with two shape
adaptive mechanisms: A multi-finger mechanism to align objects and a granular
jamming gripper to firmly hold objects. 

For the work that implemented similar functions, Bunis et al.
\cite{bunis2018caging} designed a formationally similar three-finger gripper to
cage and hold objects. The gripper has 1DoF. Together with their caging
algorithms, the gripper can cage grasp with noisy poses and grasp it to a
precise goal position. Zhang et al. \cite{zhang2002gripper} presented a gripper
with reconfigurable jaw tips and a sequencer to align objects. The work is
inspired by a previous work of Goldberg \cite{goldberg1993orienting} which used
a gripper with flat finger pads and some carefully designed work strategies for
object alignment. Mason et al.
\cite{mason2012autonomous} presented a picking system with simple grippers. The
system can predicate the pose of a grasped objects by using some pre-learned
knowledge. Wan et al. \cite{wan2018regrasp} developed a system that reorients
an object from one placement to another by using a sequence of pick-ups and
place-downs. Following the work, Cao et al. \cite{cao2016analyzing} used a
vertical pin as the intermediate location for regrasping. Ma et al.
\cite{ma2018regrasp} developed a dynamic simulator to generate stable poses and
reorient objects by using them as intermediate states. Zhou et al.
\cite{zhou2017probabilistic} presented a probabilistic algorithm that generated
sequential actions to iteratively reduce the uncertainty of the objects. Dobashi
et al. \cite{dobashi2014robust} used parallel grippers and grasp planners to align
and prepare objects for assembly. The grippers used in these work are mostly
simple parallel grippers with no special design. The working strategies helped
to meet the functional requirements.

There are also designs that use sensors and feedback control to implement
the desired functions. For example, Chen et al. \cite{chen2012hand} designed a
parallel gripper with embedded vision and force feedback to adjust poses and
perform insertion tasks. Golan et al. \cite{golan2018object} developed a swivel
mechanism to sense and position the surface of target objects. It is also an
example of positioning using sensors.

Our paper biases on the design, and relies on the geometric constraints of the
design and simple work strategies to implement the goal functions.
Different from previous studies, our gripper comprises two simple grippers. The
finger dimensions and shapes of the grippers, as well as their transmission
mechanisms, are carefully formed to meet the required functions of aligning,
picking, and arrangements. The working strategies for the
gripper to eliminate position errors and arrange screws are implemented by
taking advantages of the geometric constraints of finger shapes and gravity.

\section{Details of the Design}

\begin{figure}[!htbp]
\begin{center}
\includegraphics[width=0.45\textwidth]{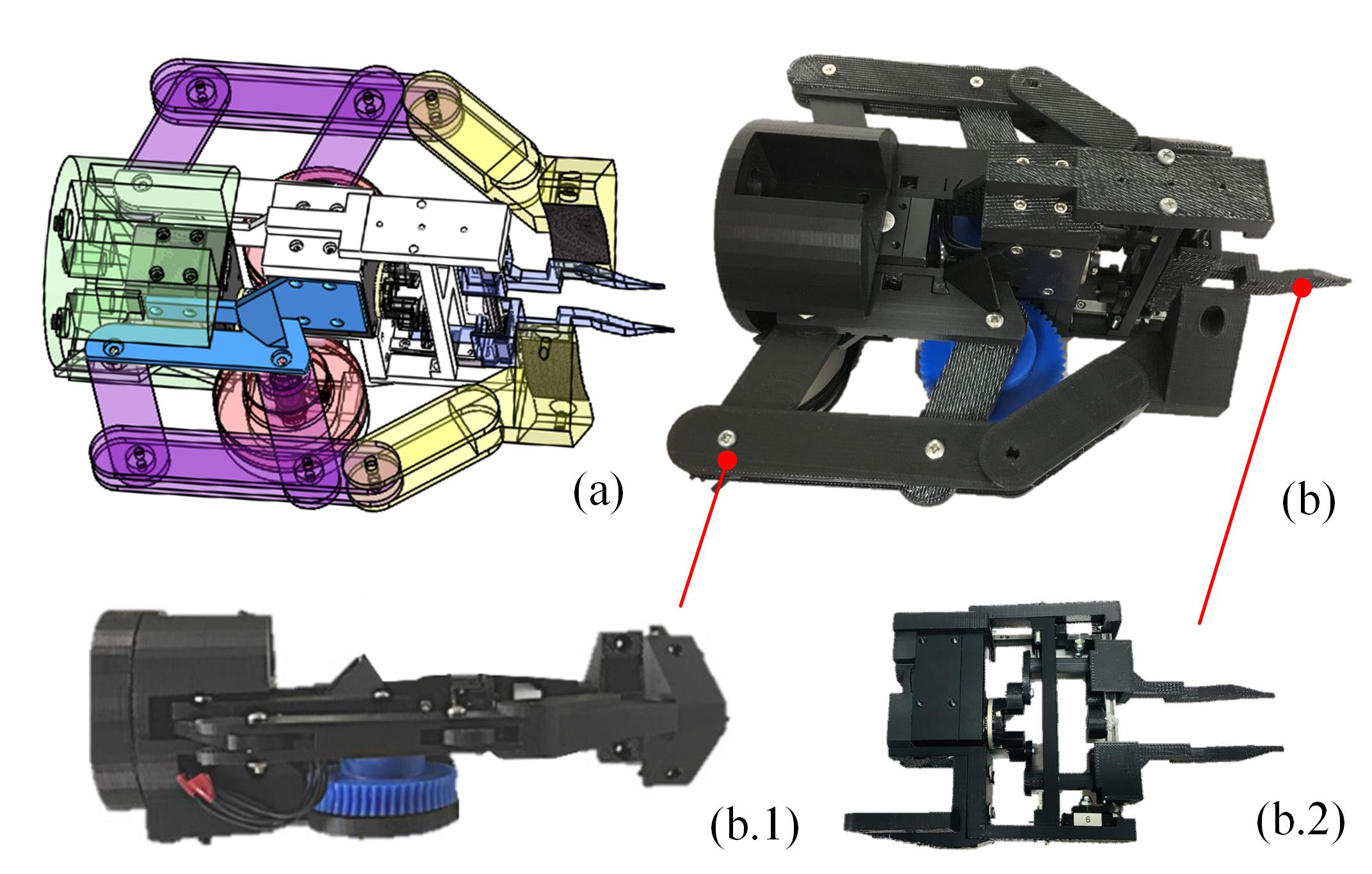}
\caption{A prototype of the proposed design.
The sketch in Fig.\ref{teaser} is repeated in (a) for comparison.
(b) An overview of the whole hand. (b.1) The outer gripper. (b.2) The inner gripper.}
\label{prototype}
\end{center}
\end{figure}

Fig.\ref{prototype} shows a prototype of the proposed design. 
The two grippers are perpendicular with each other. They have different purposes
and are driven by different mechanisms. This section discusses the design
details of the two grippers.

\subsection{The inner gripper}

\begin{figure}[!htbp]
\begin{center}
\includegraphics[width=0.45\textwidth]{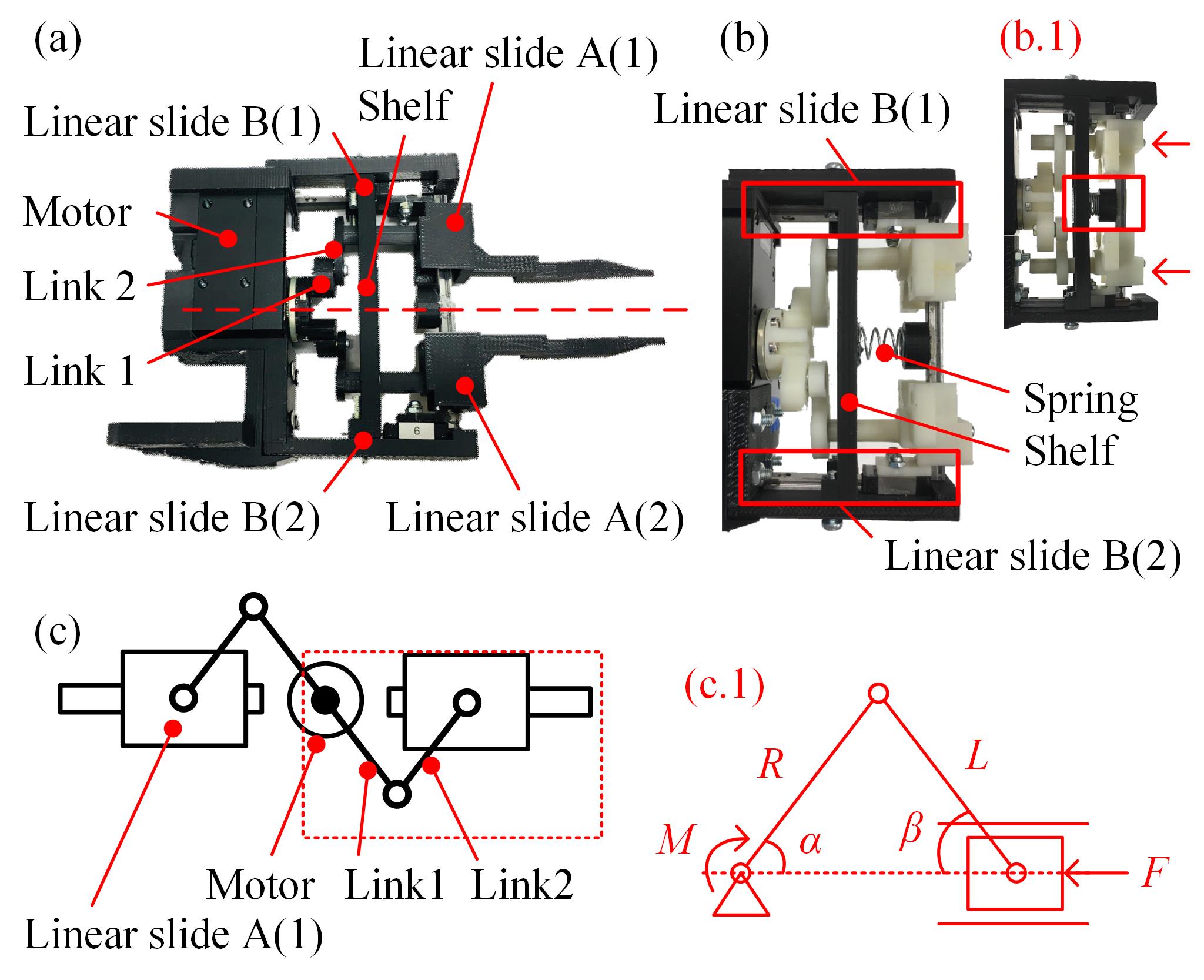}
\caption{The mechanism of the inner gripper. (a) The two fingers of the inner
gripper are mounted on linear slides A(1) and A(2). They are driven by a
slider-crank mechanism composed by Link 1 and Link 2. (b) A zoomed up view of
the compliant mechanism. When external forces are exerted at the fingertips,
the fingers will retract along the linear slide B, and the spring is compressed
(see (b.1)). When the external forces disappear, the spring recovers and moves
the fingers back to initial positions. (c) A sketch of the slider-crank
mechanism.
The definitions of various symbols are shown in (c.1):
The length of the crank is R and the length of the link which connects the
finger and crank is L. The motor angle is $\alpha$. The slider angle is
$\beta$.}
\label{crank}
\end{center}
\end{figure}

The details of the inner gripper are shown in Fig.\ref{crank}(a).
Two fingers, mounted on a linear slide A, are driven by a Dynamixel XM430-W350
motor through a slider-crank mechanism (Fig.\ref{crank}(c)). With the help of
this mechanism, the two fingers could move in parallel with each other.

A compliant mechanism, as shown in Fig.\ref{crank}(b), is introduced to
avoid hard conflicts with objects. Here, the slide A is guided by two
horizontal linear slides B(1) and B(2) so that the fingers could retreat in the
approaching direction in the presence of an resistance force. A shelf is
installed between the linear slide A and the motor and a spring is attached to
the shelf to help the fingers return to the initial position automatically when
resistance forces disappear.

The joint between joint 1 and the slider is a peg-and-hole mechanism. The
peg-and-hole mechanism secures the rotational motion between the two links.
Meanwhile, it enables linear motion in the insertion direction, and allows the
retraction of the fingers with the linear slide B. In contrast, the joint
between Link 1 and Link 2 is connected using a bearing, since only rotational
motion between the two links is needed.

\begin{figure}[!htbp]
\begin{center}
\includegraphics[width=0.49\textwidth]{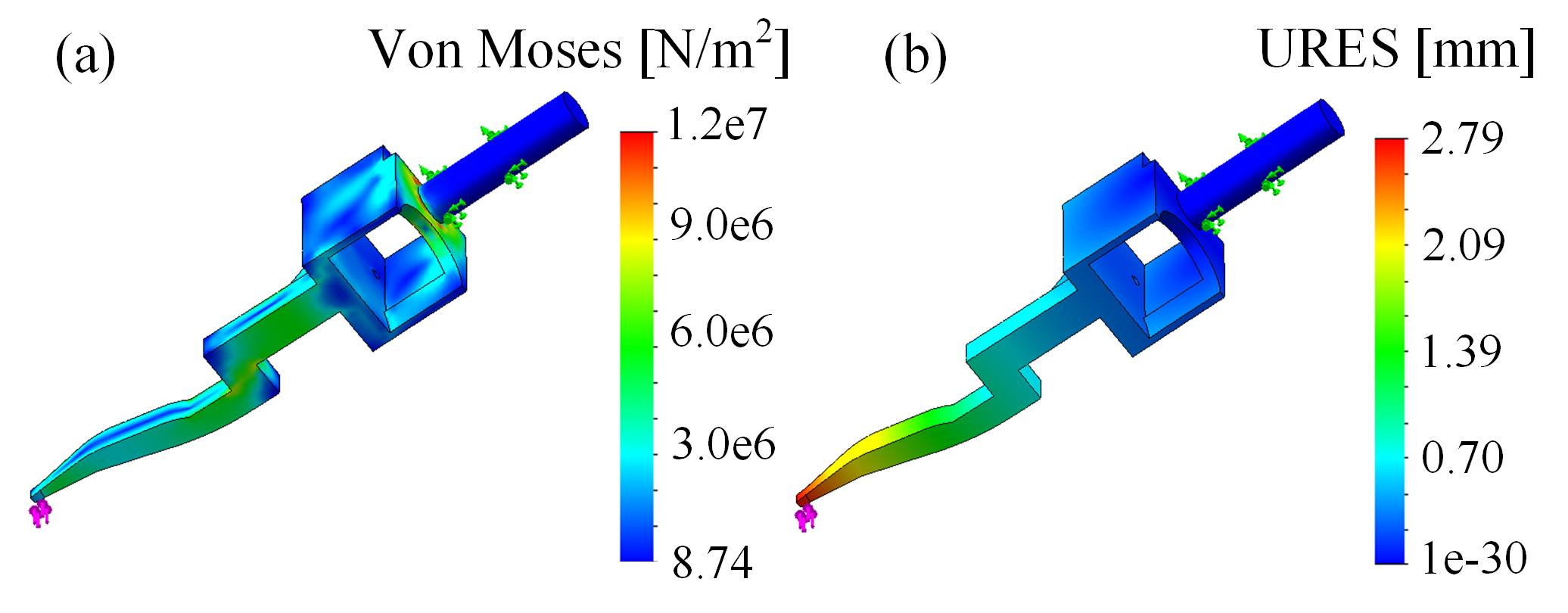}
\caption{The stiffness of the shape under a 3N lateral force applied to the
fingertip. (a) The Moses Stress distribution. (b) The deformation.}
\label{fea}
\end{center}
\end{figure}

The shape of the positioning fingers is specially designed for arranging
screws. The details will be discussed in the next Section. Also,
the fingertips are cut into a tweezers shape for picking small objects and inserting
into holes. The stiffness of the shape is validated by FEA (Finite Element
Analysis). The results are shown in Fig.\ref{fea}. Here, the material is set to
ABS. The cylindrical part of the shape is fixed as the boundary condition. As
revealed by the results, the design is safe and has an acceptable deformation
in the presence of a 3N force lateral force applied to the tips. The affordable
force increases if the fingers were made by aluminum or steel.

The relation between the motor angle and the travel of a finger is formulated
as:
\begin{equation}
	S = R+L-Rcos\alpha-Lcos\beta 
\end{equation}
In the prototype, $ R\approx L $, thus the relation between the motor
angle and the travel of a finger is equal to:
\begin{equation}
	S = 2R(1-cos\alpha)
\end{equation}
The relation between the motor torque and the force exerted by a finger is:
\begin{equation}
M = \frac{F}{cos\alpha} Rsin2\alpha
\end{equation}
The desired motor torque could be computed by the equation.

\subsection{The outer gripper}

\begin{figure}[!htbp]
\begin{center}
\includegraphics[width=0.45\textwidth]{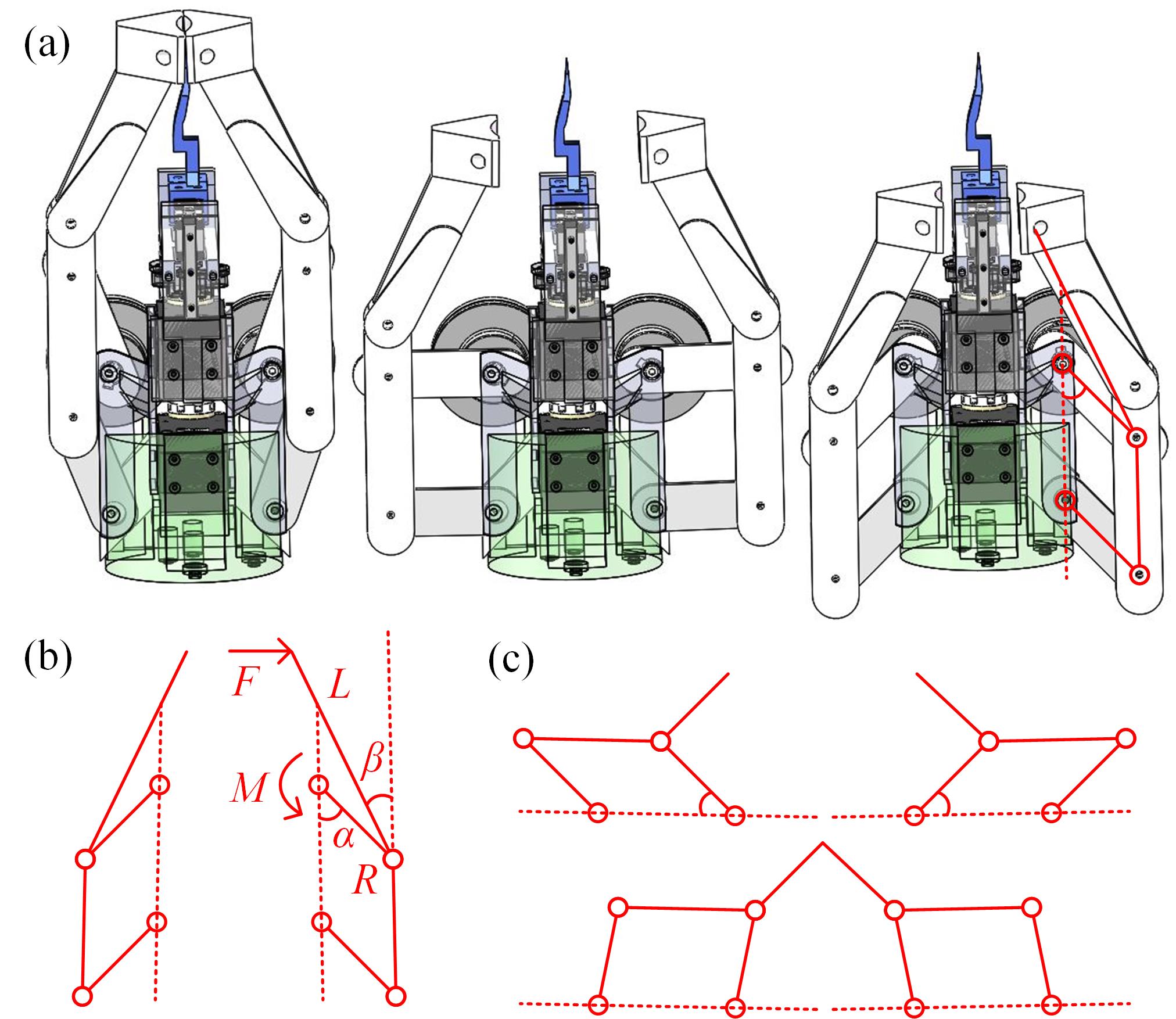}
\caption{The mechanism of the outer gripper. (a) The holding fingers are folded
back as the gripper opens. (b) The kinematic structure of the horizontally
arranged parallelogram. (c) An example of a vertically arranged
parallelogram. The vertically arranged parallelogram cannot fold the fingers
back.}
\label{para}
\end{center}
\end{figure}

The outer gripper consists of two holding fingers, which are connected to a
parallelogram driven by a motor and two worm gears. The parallelogram is
arranged horizontally to allow the holding fingers fold back and avoid conflicts
with the positioning gripper. Fig.\ref{para}(a) and (b) respectively show the
motion of the outer gripper and the kinematic structure of the horizontally arranged parallelogram. In
contrast to Fig.\ref{para}(b), Fig.\ref{para}(c) shows a vertically arranged
parallelogram. The vertically arranged one cannot fold the fingers back.

The relation between the motor angle and the travel of a finger is formulated
as:
\begin{equation}
	S = R-Rcos\alpha 
\end{equation}
The relation between the motor torque and the force exerted by a finger is:
\begin{equation}
M = F(Lcos\beta-Rsin\alpha)
\end{equation}

\begin{figure}[!htbp]
\begin{center}
\includegraphics[width=0.45\textwidth]{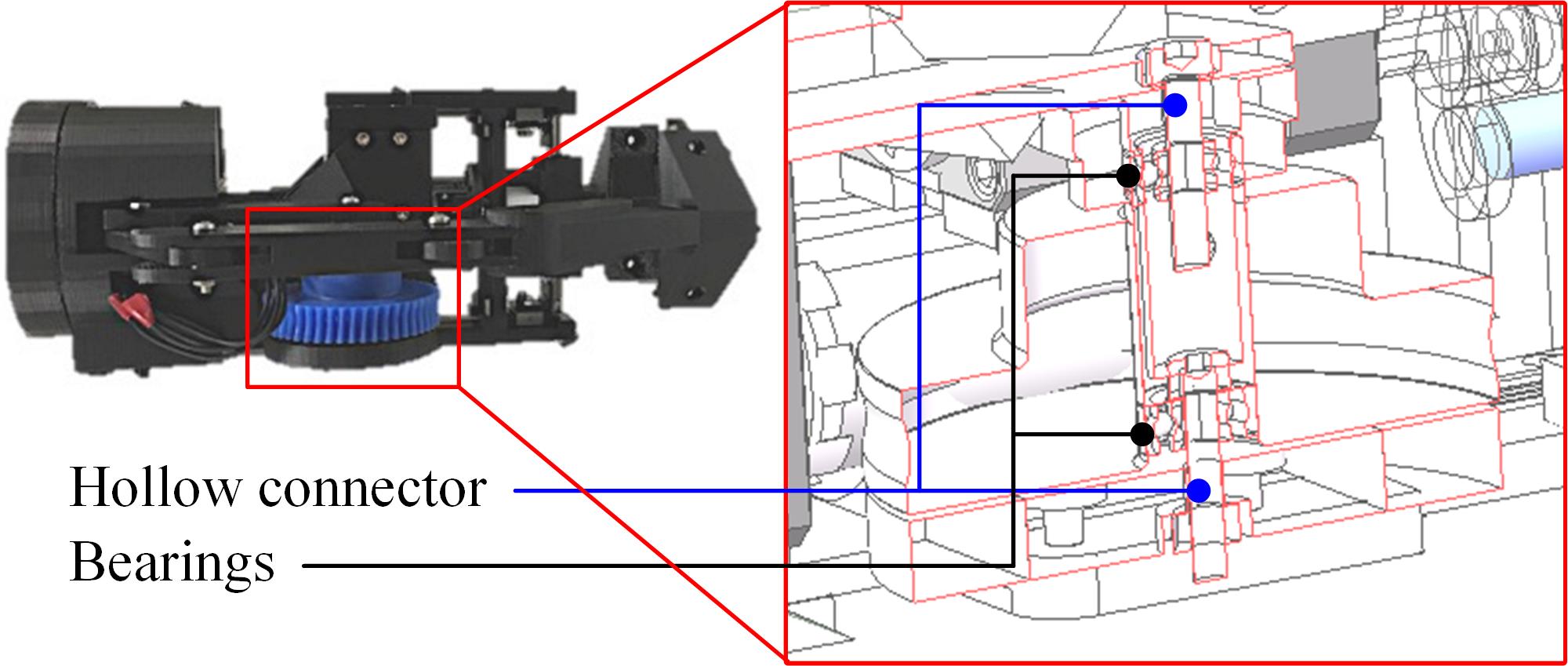}
\caption{The worm gears used to drive the outer gripper. The links connecting
the worm gears are made hollow to increase the resistance to torsional forces.}
\label{wormgear}
\end{center}
\end{figure}

The worm gears are shown in Fig.\ref{wormgear}. The links
connecting the worm gears are made hollow to increase the resistance to
torsional forces.
The motor motion is augmented and transmitted to the motion of the holding fingers
through the worm gears.

\section{Working Strategies}
\label{secws}

Strategies are designed to for the hand to align and pick up cylindrical objects
with holes, cylindrical objects without holes, and arrange screws using its
geometric constraints and gravity.


\subsection{Cylindrical objects with holes}

\begin{figure}[!htbp]
\begin{center}
\includegraphics[width=0.45\textwidth]{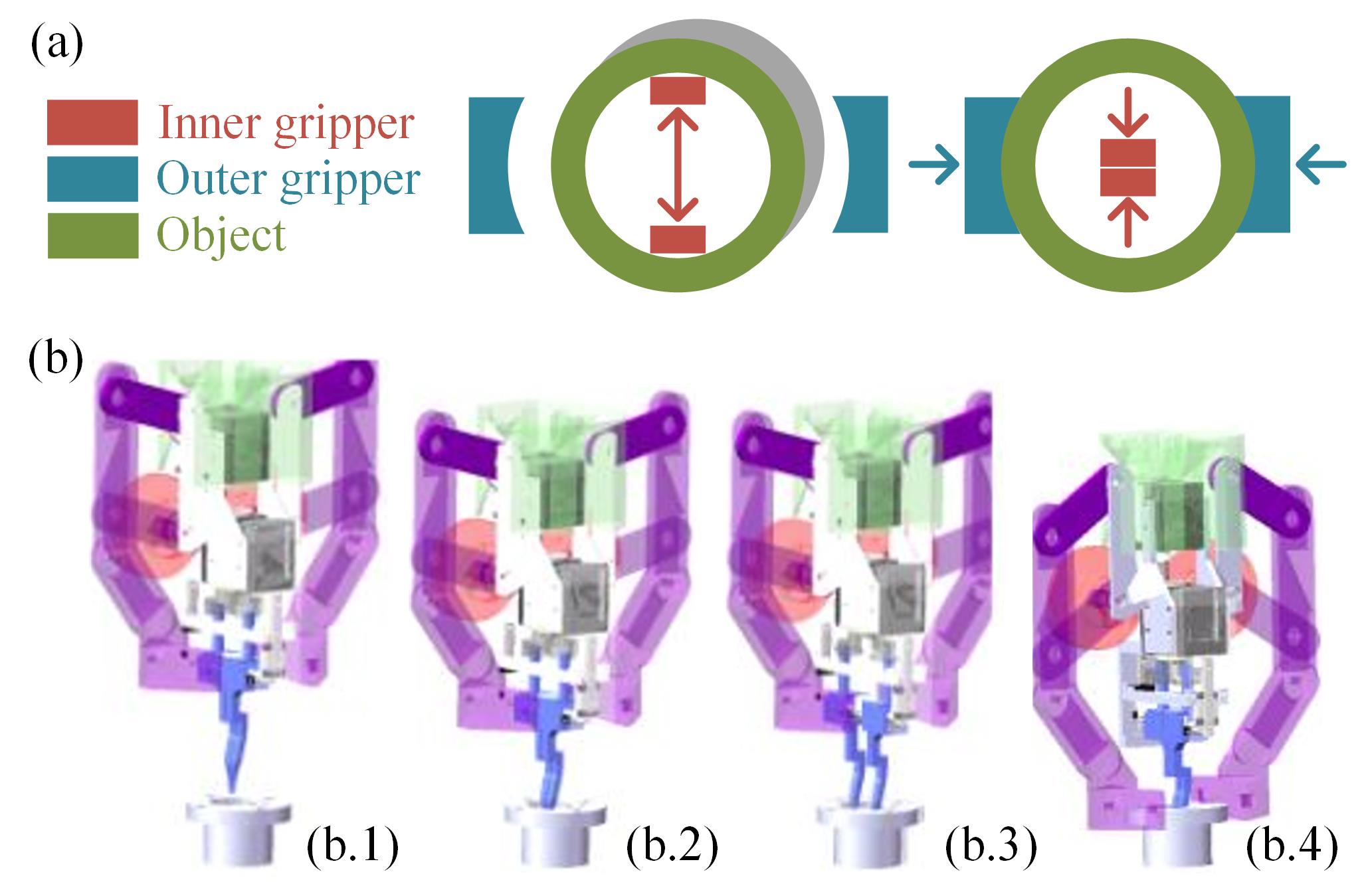}
\caption{The working strategy used to align an object with a hole. First,
the positioning fingers are stretched to align the object. Then, the holding
fingers close to hold the object stably. (a) A conceptual sketch. (b) An
illustration using CAD models.}
\label{owhstr}
\end{center}
\end{figure}

For cylindrical objects with holes, the working strategy is to use the
positioning fingers as a stretcher. The flow of the working strategy is
shown in Fig.\ref{owhstr}. It includes two steps. First, the fingertips of the
closed inner gripper are inserted into the holes and stretch from the inner side
to eliminate the position errors (see Fig.\ref{owhstr}(b.1-3)). Then, the holding gripper closes to
grasp the object stably (see Fig.\ref{owhstr}(b.4)).

\begin{figure}[!htbp]
\begin{center}
\includegraphics[width=0.45\textwidth]{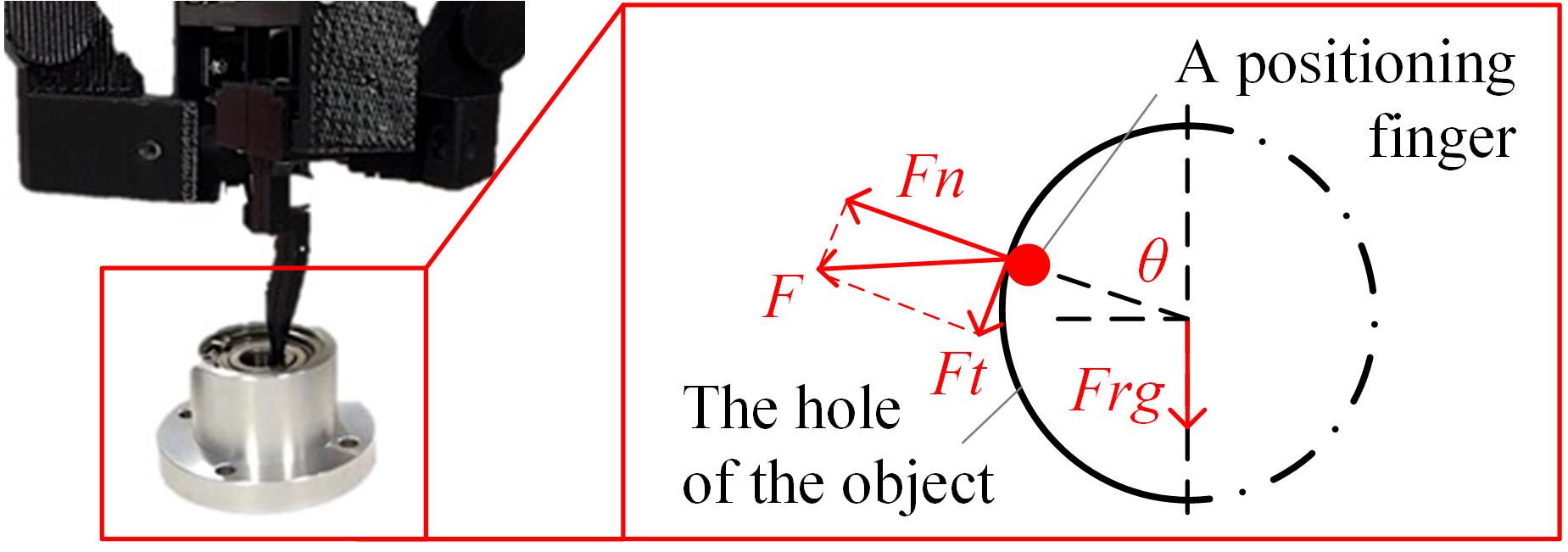}
\caption{The conditions needed to conduct a successful
alignment. The figure shows the definitions of the symbols. See
Eq.\eqref{stretchcondition} for details.}
\label{owhstrana}
\end{center}
\end{figure}

The stretching will be able to align an object under the following
condition:
\begin{equation}
2 F sin\theta cos\theta >  2 \mu_{0} F cos\theta + \mu _{og} m_{obj} g 
\label{stretchcondition}
\end{equation}
Here, $F_t = F sin\theta$ is the tangential force. $F_n = F cos\theta$ is the
normal force. They are illustrated in Fig.\ref{owhstrana}.
The friction between object and ground is defined as $F_{rg} = \mu _{og} m_{obj}
g$ where $\mu_{og}$ denotes the maximum static frictional coefficient and $m_{obj}$ denotes the mass of the
object. $\mu_{0}$ represents the maximum static friction coefficient between
the object and the positioning fingers. 

We assume an application where the mass of target objects are small and the
materials are aluminum. In that case, $\mu _{og} m_{obj} g$ can be
neglected. When $\theta$ is approaching to 90 degrees, which is our goal, $F
sin\theta cos\theta > \mu_{0} F cos\theta$ is always met and the
condition in Eq.\eqref{stretchcondition} can be satisfied with a high
probability.

After alignment, the holding fingers close to grasp target
objects stably. On the other hand, in the case of small objects with
lightweight, the positioning fingers can stably hold the objects by
stretching. The holding fingers are not used.

\subsection{Cylindrical objects without holes}

\begin{figure}[!htbp]
\begin{center}
\includegraphics[width=0.45\textwidth]{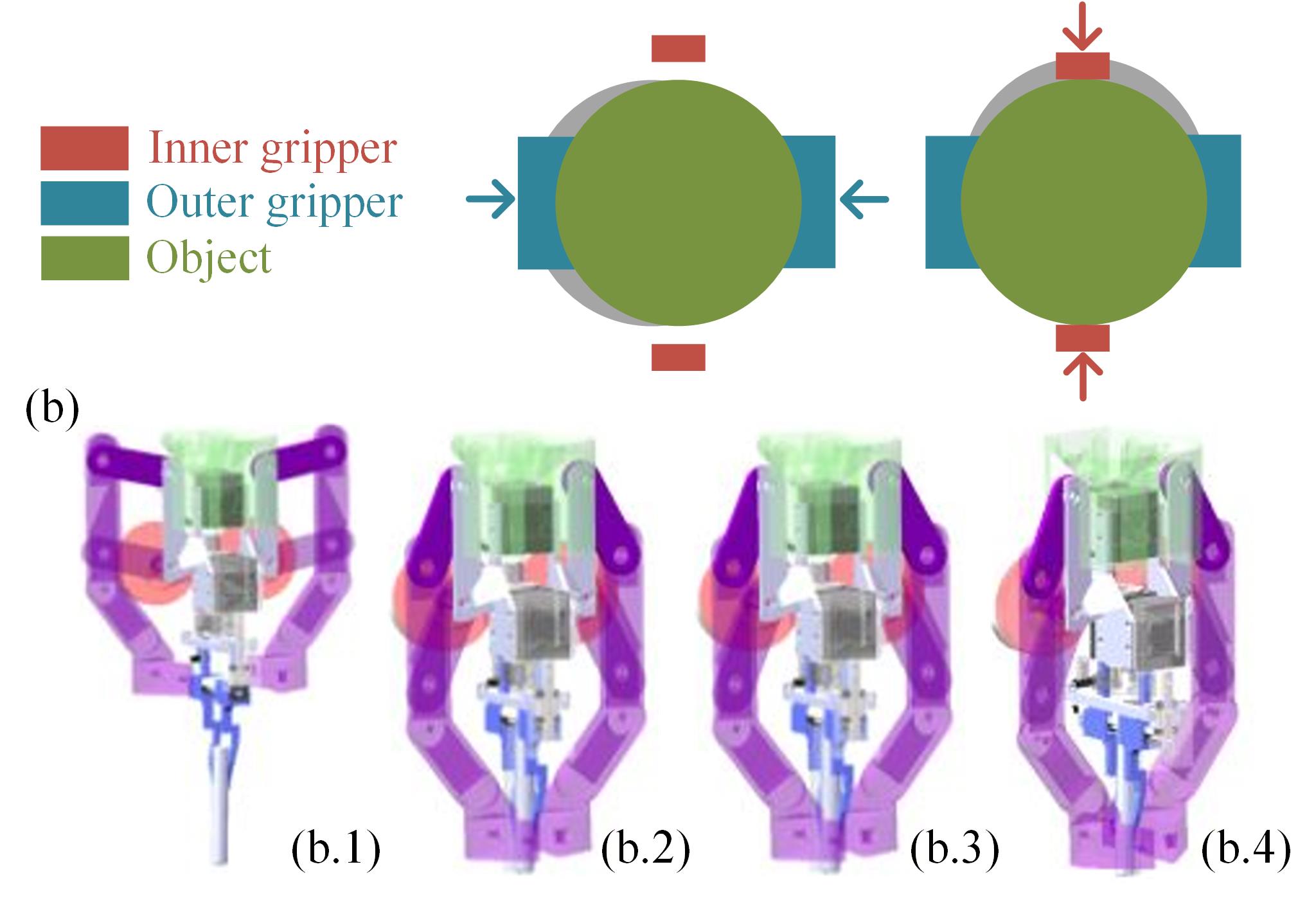}
\caption{The working strategy used to align an object without holes. Firstly,
the holding fingers close with a relatively low motor torque to eliminate
the position errors in the closing direction (see (b.2)). Then, the positioning
fingers close to eliminate the position errors of the object in a
perpendicular direction (see (b.3)). Finally, the holding fingers close firmly
to stably grasp the object (see (b.4)).}
\label{owohstr}
\end{center}
\end{figure}

For cylindrical objects without holes, 
the working strategy is to alternatively close the outer and inner grippers.
The positioning fingers are used as a squeezer, as shown in Fig.\ref{owohstr}.
First, the holding fingers close with a relatively low motor torque to eliminate
the position errors in the closing direction. 
Second, the positioning fingers close to eliminate
the position errors of the object in a perpendicular direction. The object will
be strictly aligned following the execution of the two consecutive and
perpendicular grips. Third, the holding fingers close firmly to stably grasp
the object.

\subsection{Arranging screws}

\begin{figure}[th]
\begin{center}
\includegraphics[width=0.45\textwidth]{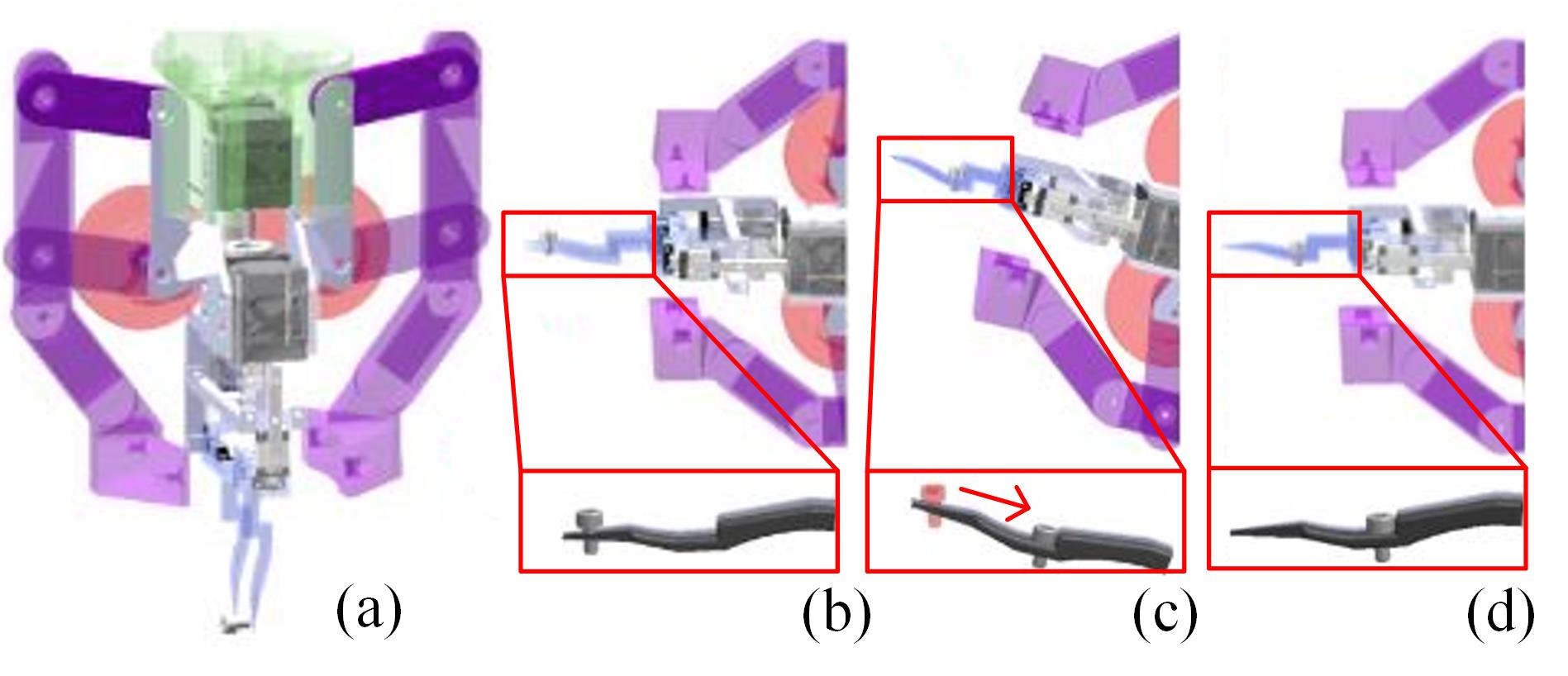}
\caption{The working strategy used to arrange screws. (a) Pick up a screw
using the positioning fingers.
(b, c) Tilt the hand with an angle, and opens the positioning
fingers with an additional distance $d$ to let the screw drop and slide.
(d) Tilt the hand back to let the screw be posed vertically at the end corner of
the positioning fingers. 
}
\label{asc}
\end{center}
\end{figure}

The screw arrangement is realized by taking advantages of the geometric
constraints as well as gravity. The working strategy of arranging screws is
shown in Fig.\ref{asc}.
The flow includes four steps: 1) The hand picks up the screw using the
positioning fingers. 2) The hand is tilted with an angle $\theta_{gri}$ to
the ground as it holds the screw. The value of $\theta_{gri}$ will be discussed
later. 3) The hand opens the positioning fingers with an additional distance $d$
so that the screw could drop and slide along the finger surface. The screw will
slide into an end corner formed by the positioning fingers after this step. 4)
The gripper is tilted back with -$\theta_{gri}$ so that the positioning fingers
are parallel to the ground. The screw will be posed vertically at the end corner
of the positioning fingers after tilting back.

\begin{figure}[th]
\begin{center}
\includegraphics[width=0.48\textwidth]{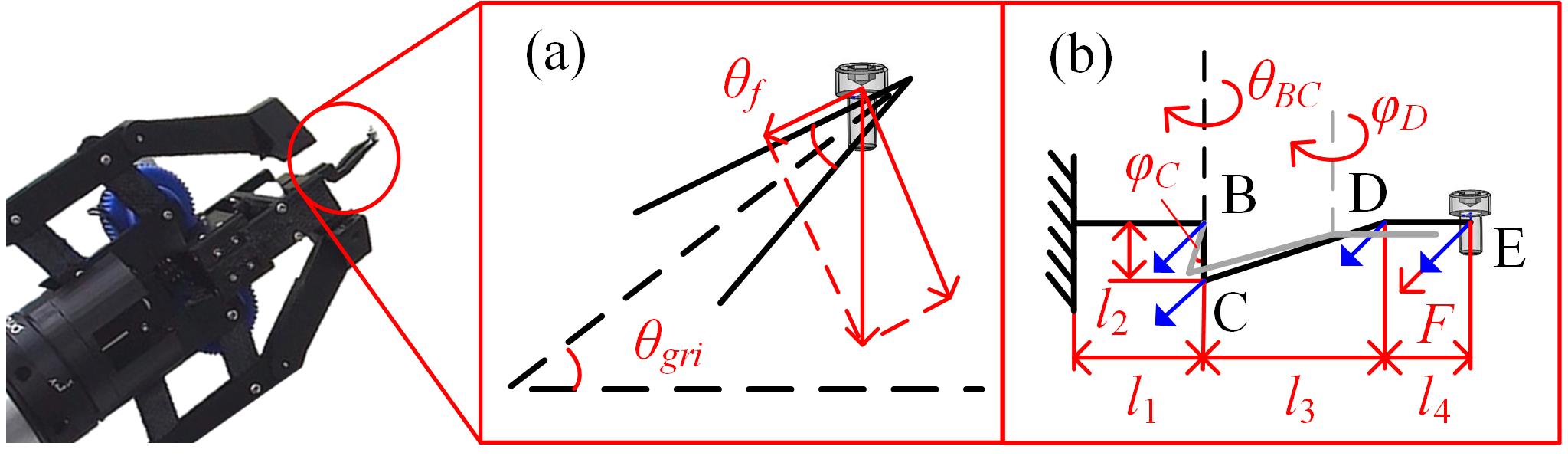}
\caption{Tilting the hand to arrange the screw. (a) Symbols used to determine
the tilting angle for successful sliding. (b) Symbols used to determine the deformation of the
fingertips and the additional opening distance.}
\label{tilt}
\end{center}
\end{figure}

In order to let the screw slide along the positioning fingers. $\theta_{gri}$
must meet the following equation.
\begin{equation}
m_{obj}gsin(\theta_{gri} - \frac{1}{2} \theta _f ) > \mu_0 cos(\theta_{gri} - \frac{1}{2} \theta _f ) 
\end{equation}
The definitions of the various
symbols are shown in Fig.\ref{tilt}(a).
$\theta _f$ indicates the interior angle of a fingertip. $m_{obj}$
denotes the mass of a screw. $\mu_0$ denotes the Coulomb friction coefficient of
the fingers. In the experiments, our target screws are M3$\times$8$mm$ and
M6$\times$12$mm$ Allen screws. $\theta_f$ is therefore designed as 20$^\circ$.

The deformation of the fingertips is taken into account to determine the
additional opening distance $d$. The finger is considered as a piece-wise beam
shown in Fig.\ref{tilt}(b). Its deformation includes a torsional deflection component and
a bending deflection component, expressed as:
\begin{equation}
	\begin{aligned}
		d_E &= d_{E_t} + d_{E_b},~where\\
		d_{E_t} &= \theta_B (l_3 + l_4),~d_{E_b} =
		\frac{F(l_3+l_4)^3}{3EI}+\frac{F(l_4)^3}{3EI}+\varphi_Cl_2+\varphi_Dl_4\\
		\theta_B & = \frac{32F(l_3+l_4)l_2}{G \pi D^4}, \varphi_C =
		\frac{F(l_2)^2}{2EI},~\varphi_D =
		\frac{F(l_4)^2}{2EI}
	\end{aligned}
\end{equation}
Here, $d_E$ indicates the deformation along the opening direction at the
fingertip E. $d_{E_t}$ is the torsional deflection component. Only BC bears the
torsional deflection. $d_{E_b}$ is the bending deflection component. $E$ is the
Young's modulus of the material. $I$ is the second moment of area. $G$ is the
modulus of rigidity of the material. $D$ is the diameter of the approximated beam at BC.
The additional opening distance $d$ must be larger than the deformation $d_E$.
Meanwhile, it should be smaller than the diameter of a screw head.
Thus, we choose $d$=$3mm$ for the two Allen screws. This value could assure
dropping and sliding, as well as avoid missing the screws. Subsequent to
releasing the positioning fingers, the screws are guided by the finger shape to
slide to an end corner.

The holding fingers are not used in screw arrangement. They are always folded
back during the task so that another tool, for instance, a vacuum fastener
could be used to pick up the screw from the top.

\section{Experiments}

The experimental section is divided into two parts. First, different
objects and various tasks are used to demonstrate the aforementioned abilities.
Second, a quantitative analysis is performed to examine the proposed design.

\subsection{Real-world tasks}

\begin{figure}[!htbp]
\begin{center}
\includegraphics[width=0.48\textwidth]{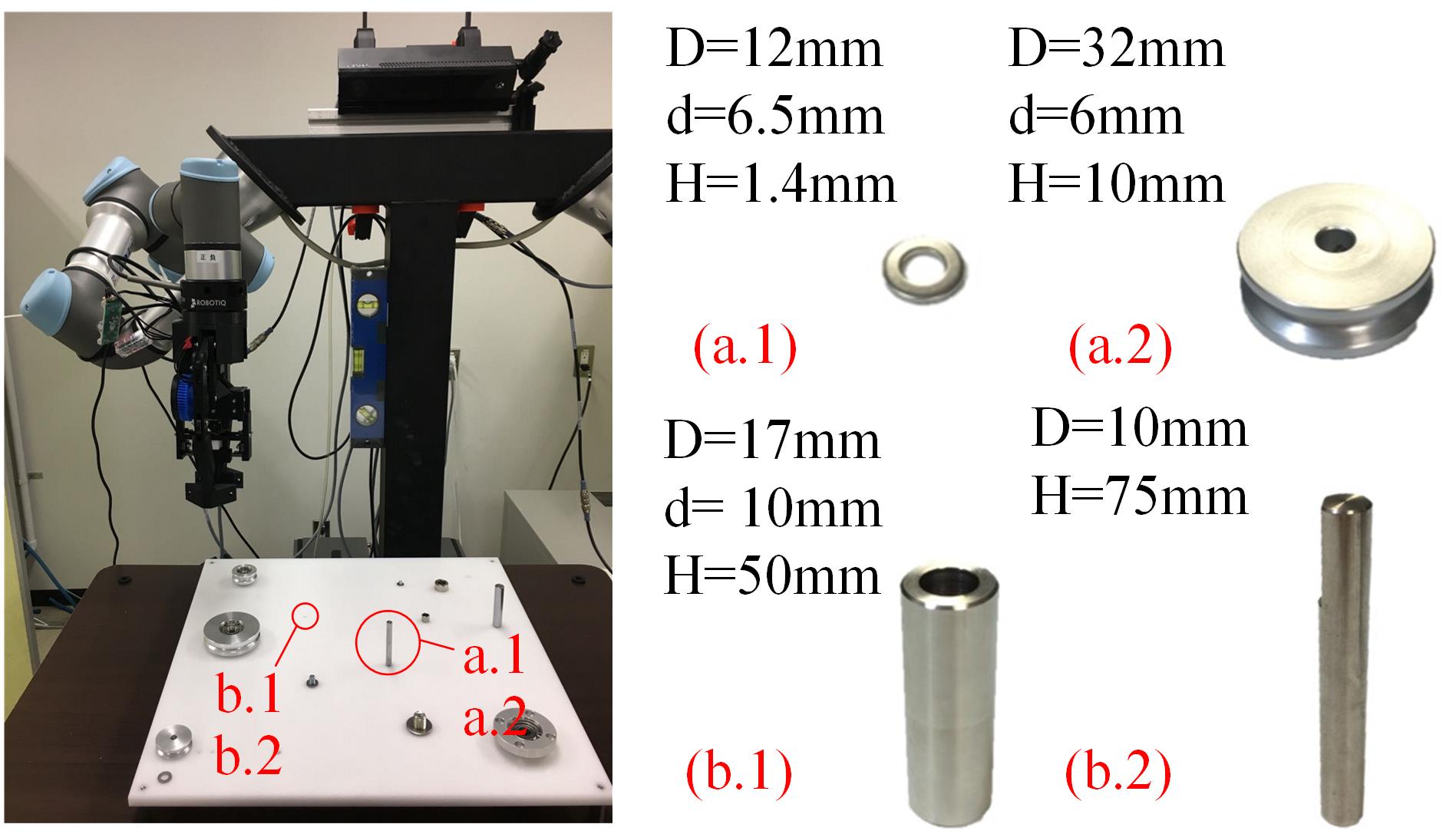}
\caption{The experimental environment includes a UR3 robot, a 3D printed
prototype of the gripper, and a task board. The task board requires attaching
four objects, a washer, a pulley, a peg with a hole, and a solid peg,
to two places on a board. The washer and the pulley will be attached to a. The
pegs will be attached to b. Meanings of the symbols: D -- The outer diameter; d
-- Diameter of the inner hole (if available); H -- The height of the object.}
\label{expenv}
\end{center}
\end{figure}

The experimental environment used for real-world executions are shown in
Fig.\ref{expenv}. The environment includes a UR3 robot, a 3D printed prototype
of the gripper, and a task board. The task board requires attaching four
objects, a washer, a pulley, a peg with a hole, and a solid peg,
to some holes and shafts on it.

\subsubsection{Objects with holes}
The peg with a hole, the washer, and the pulley, are objects with holes.
Following the work strategy introduced in Section.\ref{secws}, the hand will
stretch the inner gripper in the hole to align the object, and then use the
outer gripper to firmly grasp it. For smaller objects like the washer, the
holding fingers are not used. The positioning finger could afford enough force
to pick it up.

\begin{figure}[!htbp]
        \centering
        \begin{subfigure}{\columnwidth}
                \centering
                \includegraphics[width=.94\columnwidth]{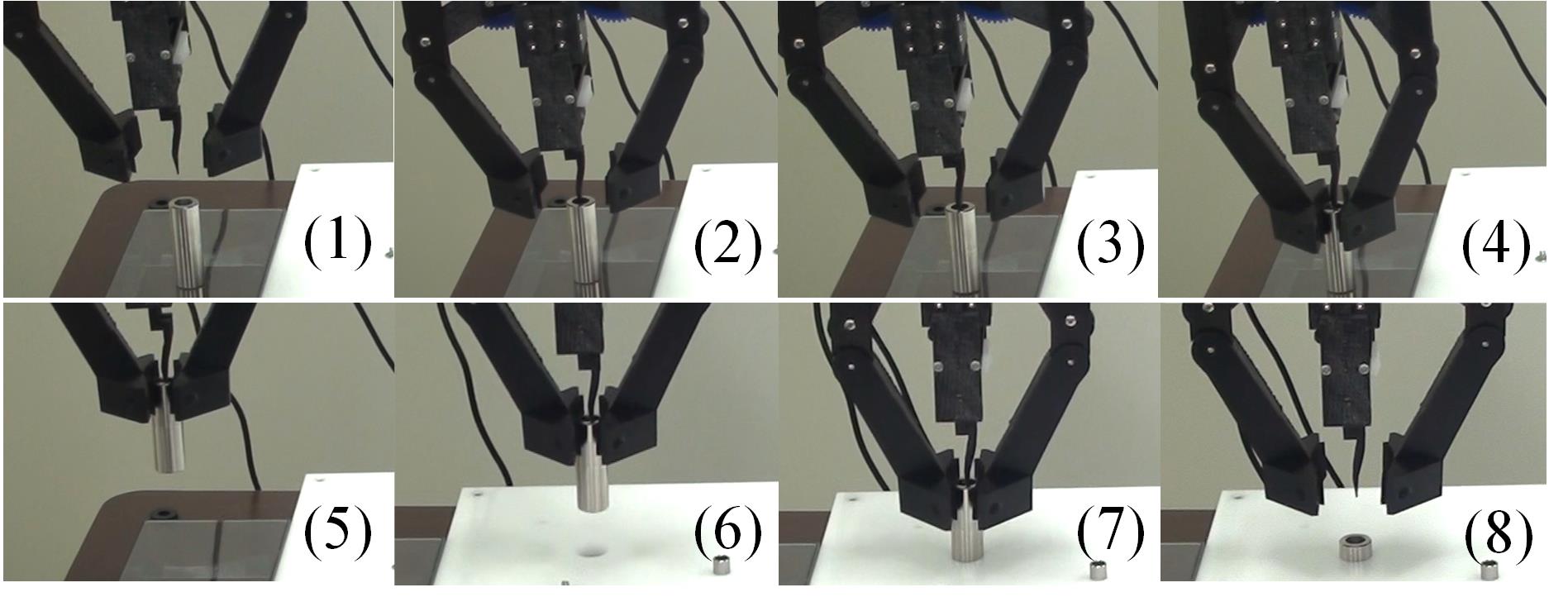}
                \caption{Attaching a peg with a hole to the task board. (1-5) Align the peg
with a hole using the positioning fingers and pick it up using the holding
fingers. (6-8) Insert the object into a hole on the task board.}
                \label{fig:intr_a}
        \vspace{0.05in}
        \end{subfigure}
        ~
        \begin{subfigure}{\columnwidth}
                \centering
                \includegraphics[width=.94\columnwidth]{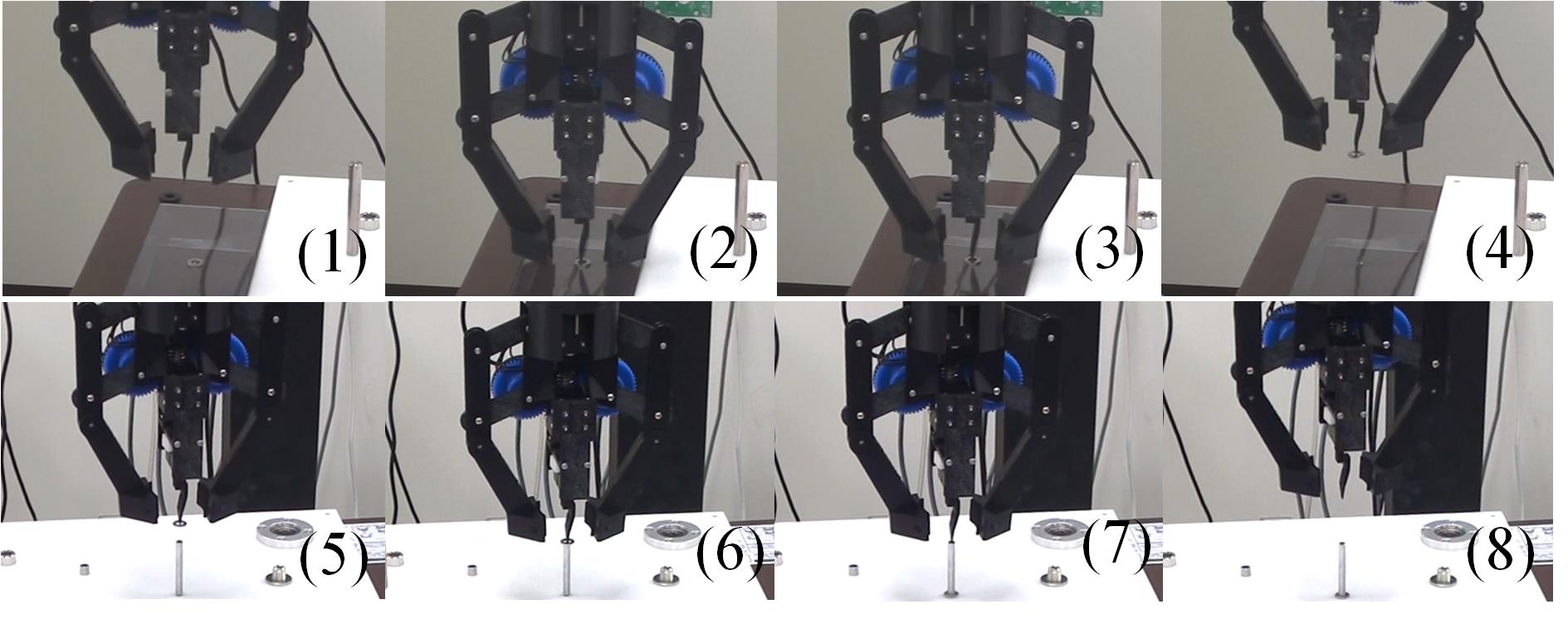}
                \caption{Attaching a washer to the task board. (1-4) Align the washer and pick
it using the positioning fingers. The holding fingers are not used. (5-8) Put
the object on a peg on the task board.}
                \label{fig:intr_b}
        \vspace{0.05in}
        \end{subfigure}
        ~
        \begin{subfigure}{\columnwidth}
                \centering
                \includegraphics[width=.94\columnwidth]{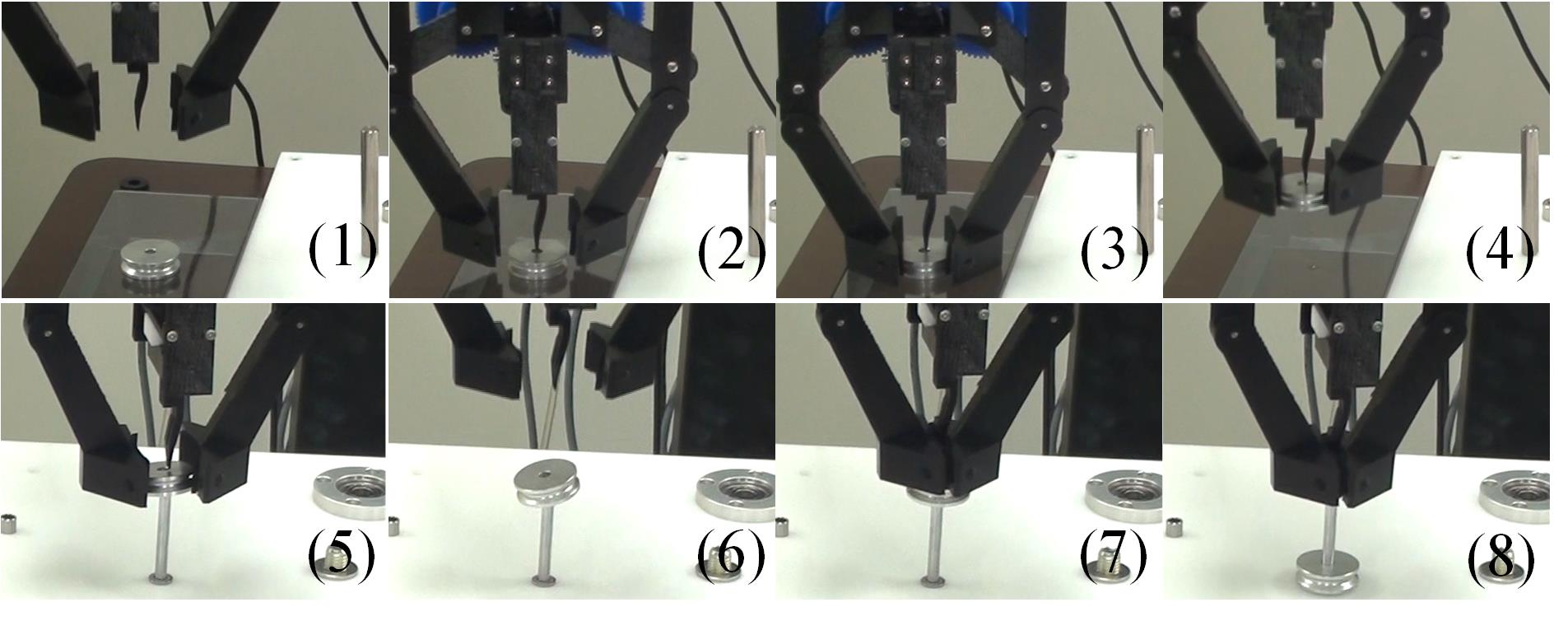}
                \caption{Attaching a pulley to the task board. (1-4) Align the pulley using the
positioning fingers and pick it up using the holding fingers. (5-6) Put the
pulley on a peg on the task board. (7-8) In case the pulley is stuck, the
holding fingers are used to push the pulley out of the stuck state.}
                \label{fig:intr_c}
        \end{subfigure}
        \caption{Manipulation sequences of the objects in
        Fig.\ref{expenv}.}
        \label{expenvres}
\end{figure}

Fig.\ref{expenvres} shows the manipulation sequences of these four
objects. Fig.\ref{expenvres}(a) is the sequence of a peg with a hole. The
object is aligned in Fig.\ref{expenvres}(a)(2-3) and picked up in
Fig.\ref{expenvres}(a)(4-5). It is inserted into a hole on a task board in
Fig.\ref{expenvres}(a)(6-8). With the help of the positioning fingers, the object
could be successfully inserted into the hole without any visual feedback.
Fig.\ref{expenvres}(b) shows the sequence of a washer. The washer is aligned and
picked up in Fig.\ref{expenvres}(b)(1-4), and is attached to a peg on the task board in
Fig.\ref{expenvres}(b)(5-8). Fig.\ref{expenvres}(c) shows the sequence of a
pulley. The pulley is aligned by the positioning fingers in
Fig.\ref{expenvres}(c)(1-2), is held and picked up in Fig.\ref{expenvres}(c)(3-4), and is
attached to a peg in Fig.\ref{expenvres}(c)(5-6). An exception we found in the
execution is large objects like the pulley tend to get stuck at a tilted pose.
In that case, the holding fingers are used to push the object out of the stuck
state, as is shown in Fig.\ref{expenvres}(c)(7-8).

%
%

\subsubsection{Objects without holes}

The solid peg is the only object that does not have holes. Following
the previously mentioned work strategy, the hand will close the outer
gripper gently to eliminate the errors in the closing direction of the holding
fingers. Then, it closes the inner fingers to eliminate the errors in a
perpendicular direction. Finally, the holding fingers close firmly to stably
grasp the object.

\begin{figure}[ht]
\begin{center}
\includegraphics[width=0.47\textwidth]{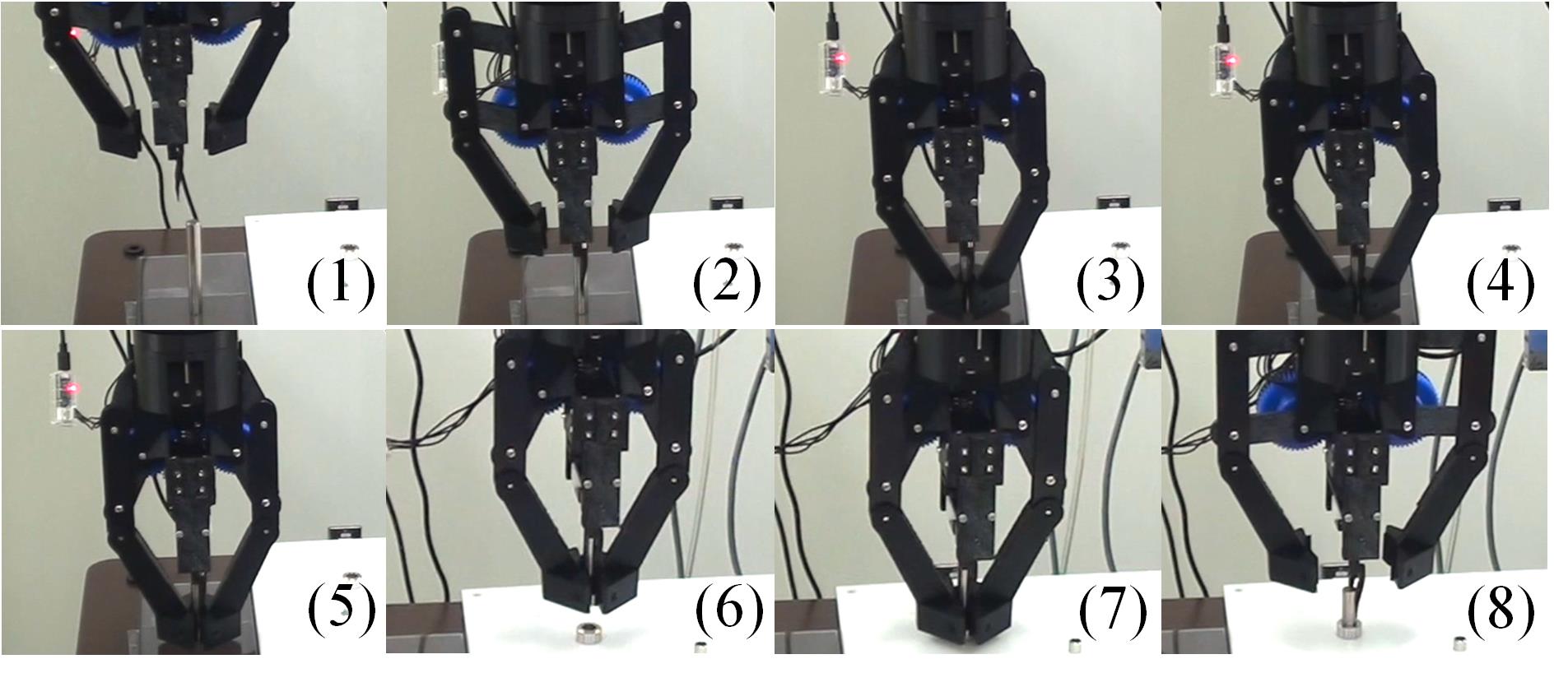}
\caption{Attaching a solid peg to the task board. (1-3) Eliminate the errors
by closing the outer gripper with a low motor torque. (4) Eliminate the errors
in a perpendicular direction by closing the inner gripper. (5) Firmly close
the outer gripper to stably grasp the peg. (6-8) Insert the peg into a hole on
the task board.}
\label{solid}
\end{center}
\end{figure}

Fig.\ref{solid} shows the execution sequence. First, the holding fingers
eliminate the errors of the peg in the closing direction in
Fig.\ref{solid}(3). Then, the positioning fingers eliminate the errors
in a perpendicular direction in Fig.\ref{solid}(4). The peg is grasped firmly by
the holding fingers in Fig.\ref{solid}(5), and is inserted into a hole on the
task board in Fig.\ref{solid}(6-8).

\subsubsection{Screw arrangement}

Besides the four objects discussed in the previous two subsections, the
same robot and hand are further used to pick up and arrange screws. Following
the work strategy presented in the Section.\ref{secws}, the gripper picks up the
screw using the positioning fingers, tilts and opens the fingers a bit to let the screw drop
and slide to an end corner, and tilts back to the horizontal state to provide an
upright screw to fastening tools like vacuum fasteners.

\begin{figure}[ht]
\begin{center}
\includegraphics[width=0.47\textwidth]{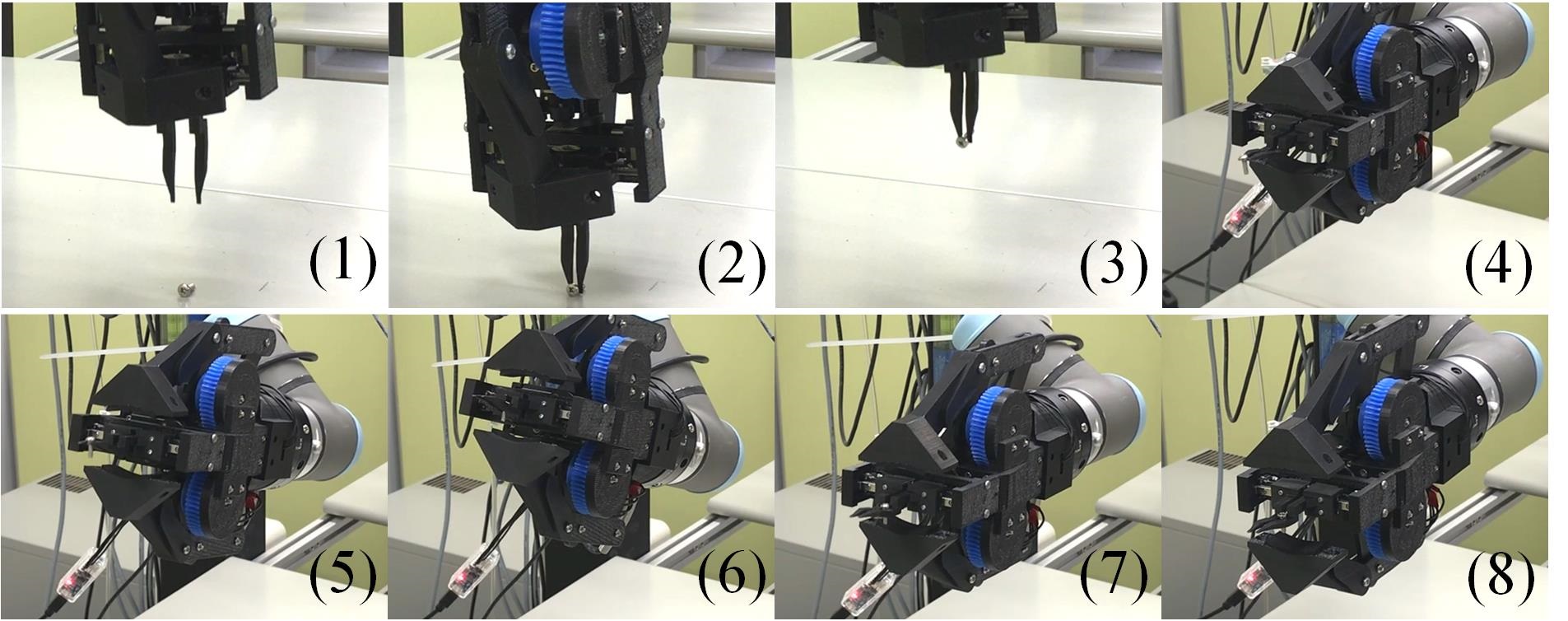}
\caption{Arranging a screw. (1-3) Pick up the screw using the
positioning fingers. (4-6) Tilt the hand and let the screw drop and
slide to an end corner of the positioning fingers. (7-8) Tilt the hand back
to let the screw pose vertically.}
\label{screw}
\end{center}
\end{figure}

Fig.\ref{screw} shows the execution sequence of screw arrangement. The screw is
picked up by the positioning fingers in Fig.\ref{screw}(1-3). The hand tilts at
Fig.\ref{screw}(4), opens a bit to let the screw drop and slide to an end corner
in Fig.\ref{screw}(5-6), and tilts back to let the screw pose vertically in
Fig.\ref{screw}(7-8).

\subsection{Quantitative analysis}

Second, quantitative analysis is performed to examine the holding force of the
outer gripper and the aligning and arranging abilities of the inner gripper.

\subsubsection{Holding force of the outer gripper}
\begin{figure}[!htbp]
\begin{center}
\includegraphics[width=0.35\textwidth]{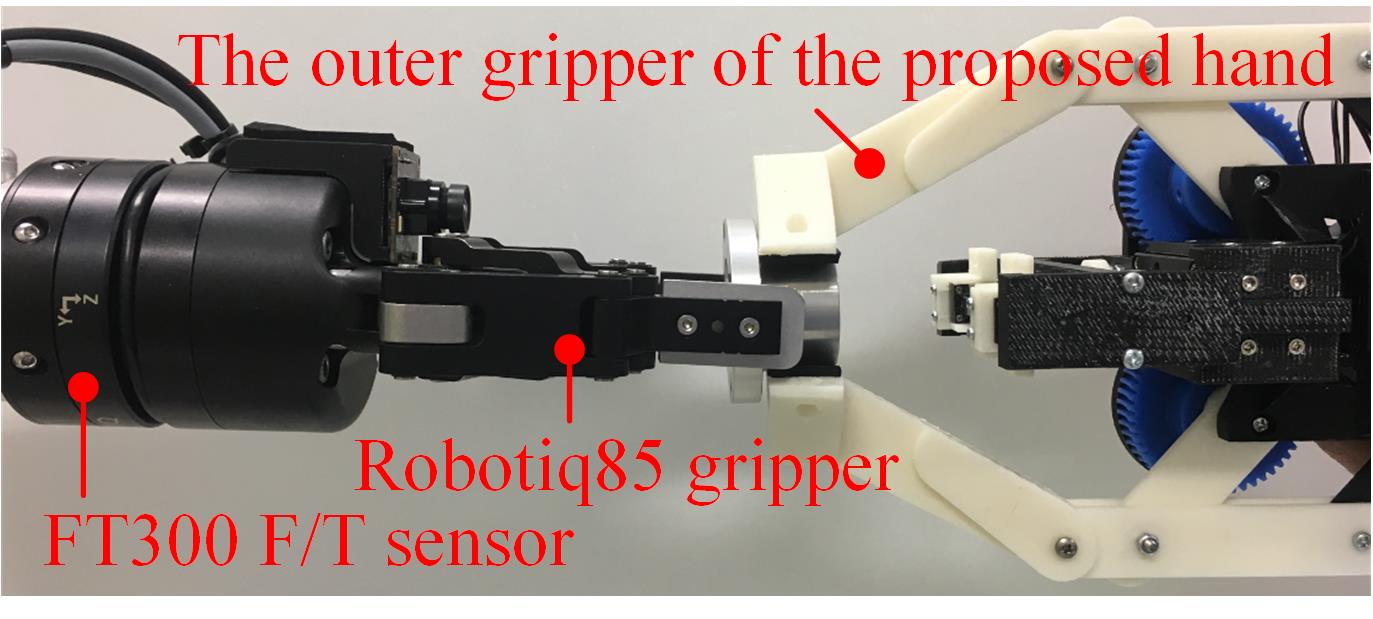}
\caption{Analyzing the holding force of the
outer gripper: A Robotiq85 gripper pulls an object out of the holding
fingers. The pulling force is measured using the FT
300 sensor.}
\label{maxforce}
\end{center}
\end{figure}

The method used to analyze the holding force of the outer gripper is shown in
Fig.\ref{maxforce}. A Robotiq85 gripper, attached to a Robotiq FT300 F/T
sensor and a UR3 robot, is used to pull an object out of the outer
gripper. The outer gripper holds one end of the object. The Robotiq85 gripper
holds the other end. The changes in forces along the pulling
direction are measured to examine the maximum holding force.

\begin{figure}[!htbp]
\begin{center}
\includegraphics[width=0.45\textwidth]{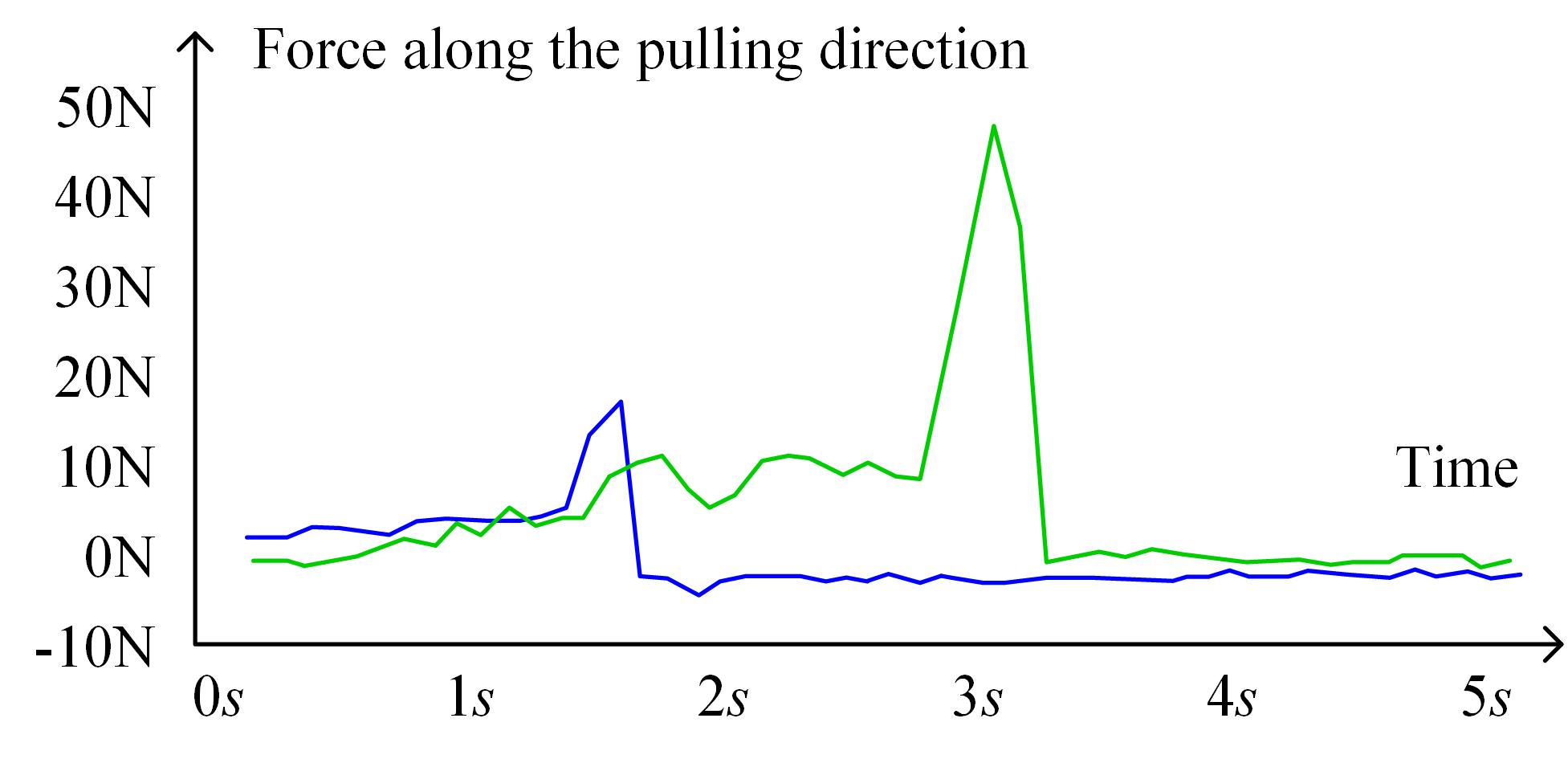}
\caption{The changes of forces along the pulling direction under 1.67\% (blue)
and 3.34\% (green) maximum toruqe.}
\label{maxforcecurve}
\end{center}
\end{figure}

Fig.\ref{maxforcecurve} shows the changes of forces under two different torques
exerted by the motor of the outer gripper.
The blue curve is the result of a 1.67\% maximum toruqe\footnote{The worm
gear has a 50:1 gear ratio. It is advisable to not use
a large percentage of the maximum torque and avoid breaking the printed
prototype.}. 
The green curve is the result of a 3.34\% maximum torque. The holding fingers
could afford around 5$kg\cdot f$ under 3.34\% maximum motor torque, which is
even larger than the maximum load of an UR3 robot (3$kg\cdot f$).

\subsubsection{Aligning ability of the positioning fingers}
In order to examine the aligning ability of the positioning fingers, we place
target objects with at erroneous positions and use the hand to pick the object and
perform peg-in-hole tasks. We measure the aligning ability of the positioning
fingers by recording the success and failure of the peg-in-hole tasks.

\begin{figure}[!htbp]
\begin{center}
\includegraphics[width=0.45\textwidth]{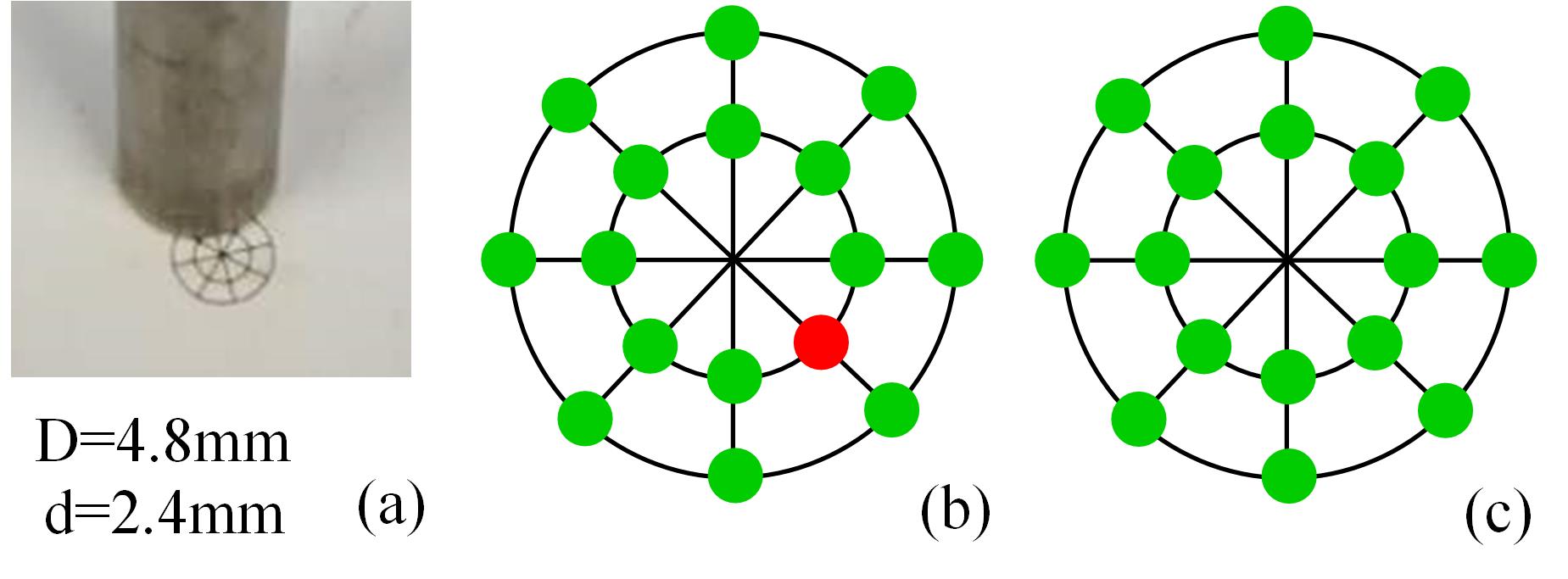}
\caption{(a) An object is placed at the 16 erroneous positions even
distributed on two concentric circles to test the aligning ability of the
positioning fingers. The outer diameter of the concentric circle is D = $4.8mm$.
The inner diameter is d = $2.4mm$. Green dot indicates the error is
eliminated. Red dot indicates a failure. (b) The
results of a peg with a hole.
(c) The results of a peg without holes.}
\label{alignab}
\end{center}
\end{figure}

The erroneous positions are manually selected as 16 evenly distributed positions on
two concentric circles, as is shown in Fig.\ref{alignab}(a). The center of the
concentric circle is the correct position. The outer diameter of the concentric
circle is D = $4.8mm$. The inner diameter is d = $2.4mm$.

Fig.\ref{alignab}(b) shows the results of an object with a hole. The dimension
of this object is shown in Fig.\ref{expenv}, 2\_1. The green dots in
Fig.\ref{alignab}(b) indicate the hand successfully finished the peg-in-hole
task when the object is placed at those erroneous positions. The errors at those
positions are eliminated by the positioning fingers. The red dot indicates a
failure. The results show that the finger may
align 2.4$mm$ position error with high success rate. The precision is satisfying
considering our low-quality 3D printer (ZORTRAX M300).
Fig.\ref{alignab}(c) shows the results of a peg without holes. All errors are
eliminated in this case. The hand has better performance in eliminating the
position errors of objects without holes.

\subsubsection{Arranging ability of the positioning fingers}
\begin{figure}[!htbp]
\begin{center}
\includegraphics[width=0.45\textwidth]{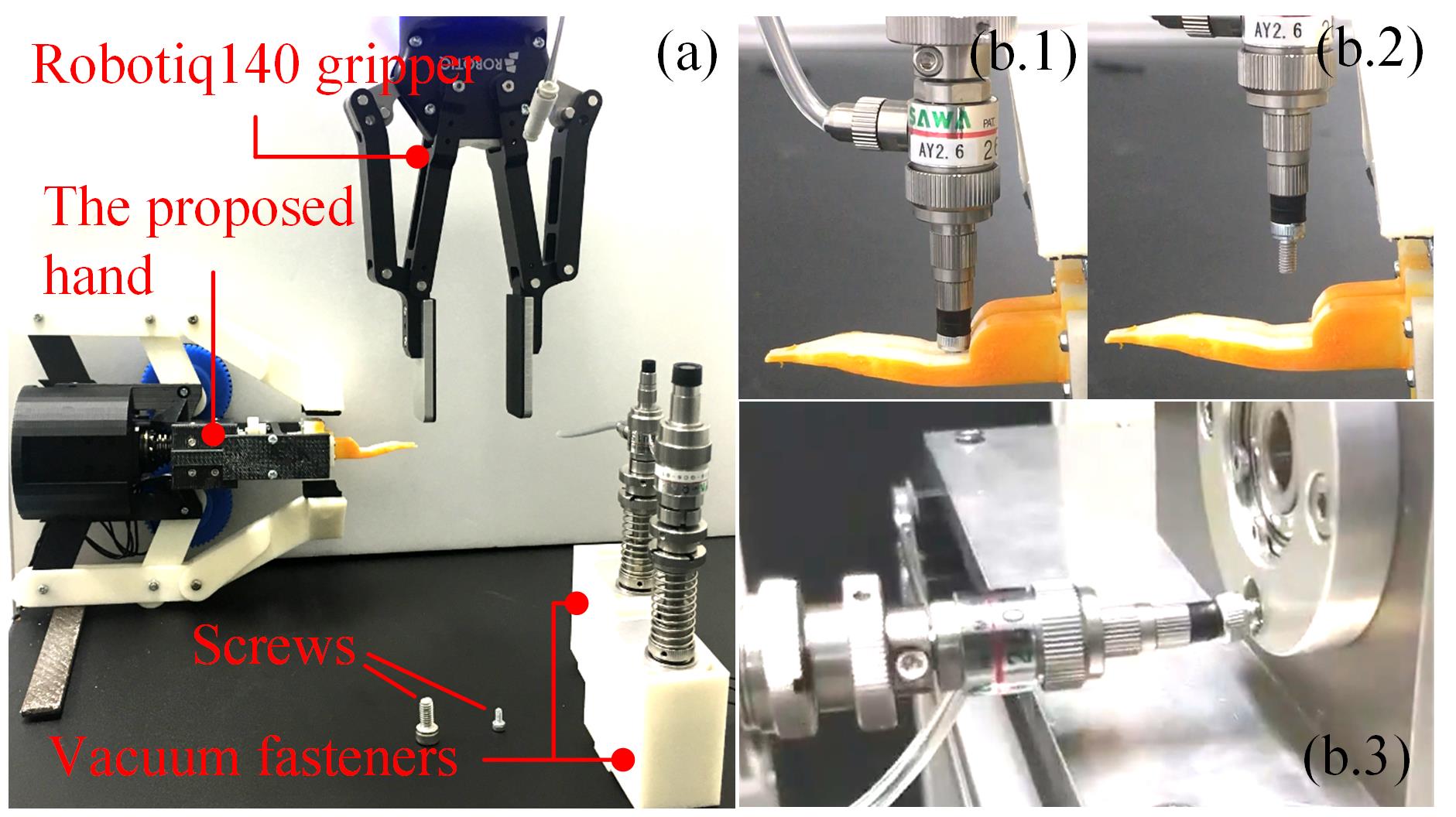}
\caption{(a) The vacuum fastener and other mechatronic devices used to test the
arranging ability of the positioning fingers. (b.1-2)
The vacuum fastener picks up a vertically arranged screw. (c) The
screw is assembled to a bearing.}
\label{arrangeab}
\end{center}
\end{figure}

To test the arranging ability of the positioning fingers, we use a vacuum
fastener shown in Fig.\ref{arrangeab}(a) to pick up the arranged screws. Two
types of screws, as is shown in the same figure, are tested. A Robotiq140
gripper is used to hold the vacuum fastener and pick up the screw from the top
after arrangement (see Fig.\ref{arrangeab}(b.1-2)). The picked screw is
assembled to a bearing housing shown in Fig.\ref{arrangeab}(b.3). In the
experiments, we didn't spot a failure. The arranging ability of the proposed
hand is highly reliable, for at least the M3 and M6 Allen screws.

\section{Conclusions and Future work}
This paper presented a novel robotic hand combining two simple grippers to
pick up and arrange objects for assembly. 
The design was validated by both quantitative analysis and
various real-world tasks. The results show that the hand has large holding force, could
eliminate position errors, could pick up and arrange small objects like washers
and screws. The hand is simple in both mechanism and control and is expected to
be used in practical systems in the future.

\bibliographystyle{IEEEtran}
\balance
\bibliography{paperKaidi}

\end{document}